\documentclass[11pt]{article}

\usepackage[preprint]{acl}

\usepackage{times}
\usepackage{latexsym}

\usepackage[T1]{fontenc}

\usepackage[utf8]{inputenc}

\usepackage{microtype}

\usepackage{inconsolata}

\usepackage{graphicx}
\usepackage{amssymb}
\usepackage{amsmath}
\usepackage{booktabs}
\usepackage{multirow}
\usepackage{tcolorbox}
\usepackage{xcolor}
\usepackage{bm}

\title{Let Relations Speak: An End-to-End LLM-GNN Soft Prompt Framework for Fraud Detection}

\author{
  \textbf{Zhixing Zuo\textsuperscript{1}},
  \textbf{Huilin He\textsuperscript{1}},
  \textbf{Jiasheng Wu\textsuperscript{1}},
  \textbf{Dawei Cheng\textsuperscript{1*}} \\
  \textsuperscript{1}School of Computer Science and Technology, Tongji University \\
  \texttt{\{2352469, huilin3\}@tongji.edu.cn}, \texttt{wujiasheng@alumni.tongji.edu.cn} \\
  {\small \textbf{* Correspondence:} \href{mailto:dcheng@tongji.edu.cn}{dcheng@tongji.edu.cn}}
}

\begin{document}
\maketitle
\begin{abstract}
In recent years, Large Language Models (LLMs) have shown great capability in processing graph tasks such as fraud detection. However, most existing methods rely heavily on rich text attributes, which poses difficulties for this domain due to the lack of textual data. Although some pioneering methods attempt to overcome it, their textualization of graph structures via hard prompts easily leads to feature distortion. Additionally, fraud detection often exhibits multi-relational complexity, where current methods struggle to capture this deep semantic information. To address these challenges, we propose LLM-GNN Soft Prompt Framework (LGSPF). Specifically, LGSPF bridges the graph structure and semantic space using soft prompt to eliminate reliance on text. 
We further introduce a parallel Graph Neural Network (GNN) encoder to translate multi-relational topologies into graph tokens for fine-grained LLM fraud comprehension.
Through end-to-end optimization, LGSPF enhances deep semantic alignment between LLM and GNN. Experiments across diverse fraud detection benchmarks demonstrate our method achieves state-of-the-art performance. Moreover, we further validate the contribution of LGSPF on enhancing the semantic interpretability of fraud behaviors.
\end{abstract}

\section{Introduction}

Fraud detection has been widely deployed to safeguard diverse real-world systems, including financial networks \citep{huang2022dgraph}, online review platforms \citep{wang2019fdgars}, cybersecurity infrastructures \citep{lo2022graphsage}, and social networks \citep{dou2020robust}. Given the nature of fraud activities, representing detection tasks as graphs has become a mainstream paradigm to capture suspicious interaction patterns \citep{ma2021comprehensive}. Recently, Large Language Models (LLMs) have exhibited remarkable reasoning and generalization abilities, driving significant progress in various graph tasks \citep{li2024survey,huang2024large}. However, deploying LLMs for fraud detection presents a unique and critical challenge: most existing LLM-based methods rely heavily on rich text attributes \citep{wang2025generalization}, while real-world fraud graphs typically contain purely numerical features due to strict privacy and security constraints \citep{LI2026129643,wang-etal-2025-llms-convert}. This leads to a Weak Text-Attributed Graph (Weak-TAG) scenario, where directly transferring existing LLM-based methods to such environment remains nontrivial.

\begin{figure}[t]
  \includegraphics[width=\columnwidth]{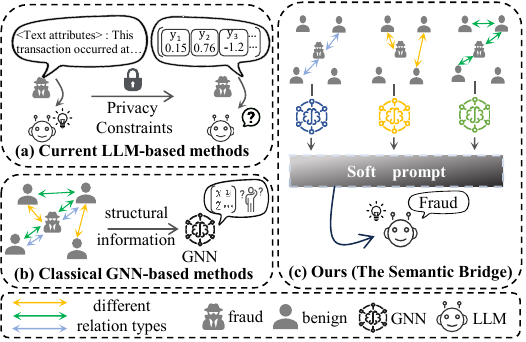}
  \caption{Comparison of different paradigms in Graph Fraud Detection: (a) Current LLM-based methods may fail in Weak-TAG scenarios. (b) Classical GNN-based methods lack interpretability on multi-relation graphs. (c) Our method seeks to bridge the semantic gap between GNNs and LLMs through soft prompt.}
  \label{fig:motivation}
  \vspace{-5pt}
\end{figure}

Notably, some pioneering methods have attempted to reduce the reliance on textual attributes, among which graph textualization via hard prompts serves as a representative approach \citep{wu2025graph,ye2024language}. By linearizing graph structures into natural language prompts, these methods offer a flexible and training-free solution that can be readily applied to various graph tasks. However, they suffer from structural-semantic misalignment. The graph topology, once flattened into natural language descriptions, loses its inherent relational semantics, leading to feature distortion. \citep{zhang2026toward,xustructure}. 
To bridge this gap, recent soft prompt \citep{li2021prefix,lester2021power} approaches, such as LLaGA \citep{chen2024llaga} and GraphTranslator \citep{zhang2024graphtranslator}, offer a promising solution by converting graph structures into LLM-compatible tokens.

However, directly applying soft prompt to encode graph into LLM for fraud detection faces challenges. In practice, fraudulent activities are rarely isolated; rather, they are highly dependent on complex, multi-relational interactions \citep{hooi2016fraudar}. To address this complexity, Graph Neural Networks (GNNs) have emerged as powerful tools that aggregate neighborhood information to capture structural patterns \citep{han2025mitigating,haghighi2024tropical}. Nevertheless, conventional GNNs essentially reduce rich relational information into static topological constraints or simplistic linear transformations, lacking interpretability for different edge types.
To overcome this, coupling multi-relational GNNs with LLMs via soft prompts appears to be a promising direction. Unfortunately, many existing LLM-GNN frameworks rely on lightweight linear projector \citep{tang2024graphgpt, cao2025instructmol} or frozen LLMs \citep{zhang2024graphtranslator}, which disconnect structural encoding from semantic reasoning — the LLM's deep semantic feedback cannot be propagated to refine graph feature learning. For fraud detection, this disconnection cripples the model's ability to exploit structural semantics for identifying suspicious behaviors.

To summarize, Figure~\ref{fig:motivation} illustrates the problem that most existing LLM-based and GNN-based methods are facing in fraud detection. It motivates us to explore the possibility in narrowing the gap between structural information and semantic reasoning, where GNNs are naturally fit in extracting topological features from multiple relations and LLMs specialize in deep semantic reasoning. We propose \textbf{LGSPF} (\textbf{L}LM-\textbf{G}NN \textbf{S}oft \textbf{P}rompt \textbf{F}ramework for Fraud Detection).
In particular, the operation pipeline of LGSPF progresses through three highly correlated stages. First, for multi-relational graph feature encoding, we implement a parallel GNN encoder to capture structural representations under multiple relations and project them into the embedding space of LLMs. Subsequently, during soft prompt and semantic mapping, we introduce multiple special tokens as interfaces and map them with simple natural-language descriptions to help LLMs realize the semantic difference between relations. Finally, we design the training procedure as an end-to-end optimization, realizing a deep-level alignment of structure and semantics. Our contributions are summarized as follows:

\begin{itemize}
    \item We propose LGSPF, an end-to-end LLM-GNN soft prompt framework for Fraud Detection, which bridges the gap between multi-relational graph structure and semantic space of LLMs under Weak-TAG scenarios.
    \item We introduce a parallel GNN encoder for feature extraction under a multi-relational graph, a soft prompt mechanism to seamlessly bridge graph structures with LLM reasoning, and an end-to-end training paradigm for deep structural-semantic alignment.
    \item Experiments conducted on three fraud detection datasets verify the effectiveness of our proposed framework. Furthermore, the results demonstrate that LGSPF delivers outstanding performance over several benchmarks and provides a new discovery in advancing graph fraud detection from black-box prediction to behavior recognition.
\end{itemize}

\begin{figure*}
    \centering
    \includegraphics[width=\textwidth]{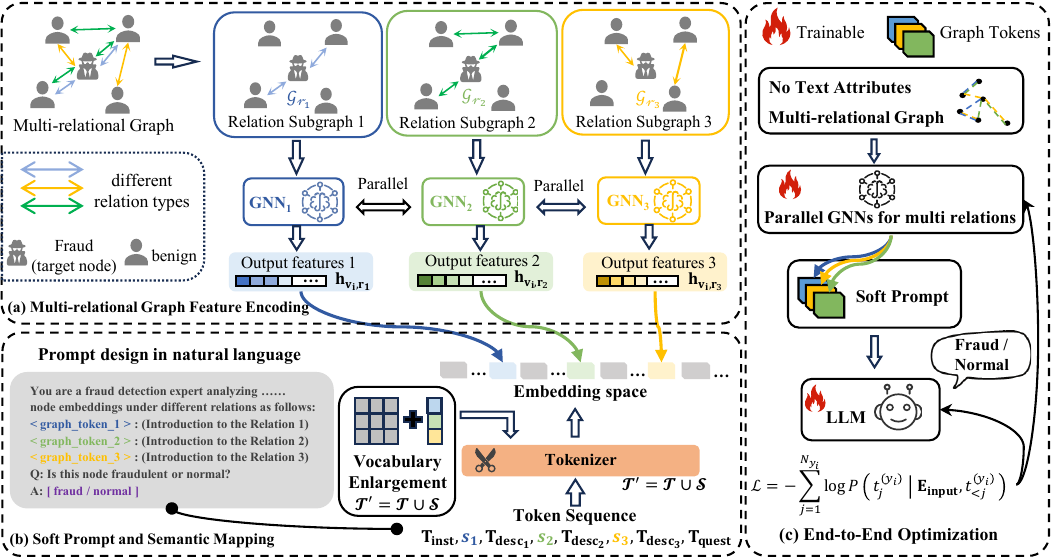}
    \caption{The overall architecture of our proposed LGSPF framework, which seamlessly aligns multi-relational graph structures with the large language model via an end-to-end soft prompting pipeline.}
    \label{fig:structure}
    \vspace{-5pt}
\end{figure*}

\section{Related Work}

\subsection{LLMs for Graph Tasks}
With the impressive abilities in generalization and reasoning, Large Language Models (LLMs) have been implemented to diverse graph tasks \citep{li2024survey}. Recent advances can be categorized into LLM-as-enhancer and LLM-as-predictor. The former paradigm, including works like GALM \citep{xie2023graph} and LLMRec \citep{wei2024llmrec}, focus on utilizing LLMs to process the text information and enhance the feature quality. LLM-as-predictor methods treats LLMs as predictors for diverse graph-related tasks such as classifications and reasonings. For instance, GraphGPT \citep{tang2024graphgpt} and GraphTranslator \citep{zhang2024graphtranslator} uses dual-stage instruction tuning to align LLMs with graph structural information, meanwhile HiGPT \citep {tang2024higpt} and DGP \citep{li2026dgp} implements instruction tuning to heterogeneous graphs. However, most of existing methods remain shackled by text attributes and lack an end-to-end tuning for deeply aligning structures and semantics, which our work addresses.

\subsection{Fraud Detection on Graph}
The topological structure of the graph provides a powerful discriminative basis for identifying fraudsters. Since the fraudulent activities exhibit complex interactive behaviors, researchers tend to construct multi-relation graphs with heterogeneous edges in distinct semantic types \citep{wu2020comprehensive,liu2018heterogeneous}. To handle this challenge, diverse Graph Neural Network (GNN) based methods have emerged, leveraging community enhancement \citep{han2025mitigating}, attention mechanisms \citep{haghighi2024tropical}, contrastive and reinforcement learning \citep{wang2023fraud,dou2020enhancing}. In contrast,  recent works such as FLAG \citep{yang2025flag} and MLED \citep{huang2025can} conduct LLMs to handle graph fraud detection, but they struggle to dynamically reason over the complex multi-relational topologies intrinsic to fraudulent behaviors. Differently, our LGSPF introduce soft prompt to bridge the GNNs and LLMs, successfully aligning the topological information in multi-relational graphs and the semantics reasoning, which significantly improve the performance and interpretability on fraud detection.

\section{Methodology}

Figure~\ref{fig:structure} illustrates the overall architecture of our proposed LGSPF. The pipeline is structured into three sections. Initially, for multi-relational graph feature encoding, we implement a parallel GNN encoder to capture relation-specific topological representations from independent subgraphs, projecting them to the embedding dimensions of the LLMs. Subsequently, during soft prompt and semantic mapping, we introduce special tokens to a unified prompt design, effectively bypassing flattened textualization. Finally, the framework employs an end-to-end optimization paradigm, where the GNN and LLM is jointly optimized to align the structural information and semantic reasoning.

\subsection{Multi-relational Graph Feature Encoding}

A graph with multi relations is defined as $\mathcal{G}_{r_i} = (\mathcal{V}, \mathcal{E}, \mathcal{R})$, where $\mathcal{V} = \{v_1, \dots, v_n\}$ is the set of $n = |\mathcal{V}|$ nodes, $\mathcal{R} = \{r_1, \dots, r_m\}$ is the set of $
m$ relations. $\mathcal{E} \subseteq \mathcal{V} \times \mathcal{R} \times \mathcal{V}$ denotes the set of edges. For a specific relation $r_i \in \mathcal{R}$, we define its corresponding subgraph as $\mathcal{G}_{r_i} = (\mathcal{V}, \mathcal{E}_{r_i})$, where $\mathcal{E}_{r_i} = \{(u, r_i, v) \mid (u, r_i, v) \in \mathcal{E}\}$. Consequently, we separate the original multi-relational graph into $m$ homogeneous subgraphs, which satisfy: 
\begin{equation}
  \label{eq:1}
  \mathcal{E} = \bigcup_{i=1}^{m} \mathcal{E}_{r_i} 
\end{equation}
\begin{equation}
    \label{eq:2}
    \mathcal{E}_{r_i} \cap \mathcal{E}_{r_j} = \emptyset, \forall i \neq j \in [1, m]
\end{equation}
In the scenario of Weak-TAG, node $v_i$ has a numeric feature vector $\mathbf{x}_i \in \mathbb{R}^d$ where $d$ is the feature dimension. For the binary graph fraud detection dataset, the node $v_i$ has label $y_i \in \{0, 1\}$, where $y_i = 1$ indicates a fraudulent node and $y_i = 0$ indicates a normal node. In the semi-supervised setting, the overall node set $\mathcal{V}$ is partitioned into labeled $\mathcal{V}_{\text{L}}$ and unlabeled $\mathcal{V}_{\text{U}}$ subsets. While model optimization and evaluation are strictly confined to $\mathcal{V}_{\text{L}}$, the unlabeled nodes in $\mathcal{V}_{\text{U}}$ are preserved to provide fundamental topological information during the GNN message-passing phase.

Given a node $v_i \in \mathcal{V}$ as the detection target, our primary objective is to capture its diverse topological patterns under different relational contexts. To achieve this, we introduce a parallel GNN encoder. Specifically, we deploy an independent GNN branch for each relational subgraph $\mathcal{G}_{r_j}$ where $j \in [1, m]$. Notably, the weights are not shareable among different branches, which means that the distinct structural semantics inherent to each relation could be extracted respectively. This multi-view decoupling essentially provides fine-grained, relation-specific evidence for the subsequent cognitive reasoning using LLMs. Formally, the parallel encoding process is abstracted as:
\begin{equation}
  \label{eq:4}
  \begin{split}
    \mathbf{h}_{v_i, r_j} = & \text{GNN}_{r_j}(\mathbf{x}_i, \mathcal{G}_{r_j}; \bm{\Theta}_{r_j}) \\
    & \in \mathbb{R}^{d_{\text{emb}}}, \quad \forall j \in [1, m]
  \end{split}
\end{equation}
where $\mathbf{\Theta}_{r_j}$ denotes the learnable parameters specific to relation $r_j$. Furthermore, in order to smoothly align with the latent embedding space of LLMs, we project the output dimension of all the GNN branches to match $d_{\text{emb}}$, facilitating seamless soft prompt integration.

It should be emphasized that our parallel architecture can adapt to various mainstream neural network-based graph structural encoders. In this work, we instantiate $\text{GNN}_{r_j}$ using the highly recognized GraphSAGE \citep{hamilton2017inductive} due to its outstanding performance in sparse graph situations. However, distinct from its original mini-batch sampling design, we perform full-batch neighborhood aggregation. 
This design choice guarantees deterministic message passing and prevents the structural information loss typically caused by random neighborhood sampling. For the $l$-th layer of the encoder under relation $r_j$, the representation of node $v_i$ is updated through a two-stage paradigm:
\begin{equation}
    \label{eq:5}
    \begin{split}
    \mathbf{h}_{\mathcal{N}_{r_j}(v_i)}^{(l)} = & \; \text{AGGREGATE}^{(l)} \Big( \\
    & \big\{ \mathbf{h}_{u, r_j}^{(l-1)}, \forall u \in \mathcal{N}_{r_j}(v_i) \big\} \Big)
\end{split}
\end{equation}
\begin{equation}
   \label{eq:6}
   \mathbf{h}_{v_i, r_j}^{(l)} = \sigma \Big( \mathbf{W}_{r_j}^{(l)} \cdot \big[ \mathbf{h}_{v_i, r_j}^{(l-1)} \parallel \mathbf{h}_{\mathcal{N}_{r_j}(v_i)}^{(l)} \big] \Big)
\end{equation}
where $\mathcal{N}_{r_j}(v_i)$ represents the 1-hop neighbors of node $v_i$ in subgraph $\mathcal{G}_{r_j}$. The $\text{AGGREGATE}^{(l)}(\cdot)$ function is implemented as MEAN pooling, which computes the element-wise average of the neighbors' representations. $\sigma(\cdot)$ denotes the Exponential Linear Unit (ELU) activation function, and $\parallel$ represents the concatenation operation. $\mathbf{W}_{r_j}^{(l)}$ is the layer-specific transformation matrix. 

The initial node representation is set to the input features  $\mathbf{h}_{v_i, r_j}^{(0)} = \mathbf{x}_i$.
We conduct $K$ layers of aggregation (i.e., aggregating $K$-hop neighborhood features) to obtain the final outputs. 
Importantly, the learnable parameter set $\mathbf{\Theta}_{r_j}$ introduced in Eq.~\ref{eq:4} is defined as below:
\begin{equation}
    \label{eq:7}
    \bm{\Theta}_{r_j} = \left\{ \mathbf{W}_{r_j}^{(l)} \right\}_{l=1}^{K}.
\end{equation}
 Ultimately, for each target node $v_i$, we extract a set of $m$ relation-specific structural embeddings $\mathcal{H}_{v_i} = \{ \mathbf{h}_{v_i, r_1}^{(K)}, \dots, \mathbf{h}_{v_i, r_m}^{(K)} \}$.

\begin{table}
    \centering
    \begin{tcolorbox}[
        colback=blue!4!white, 
        colframe=blue!50!gray, 
        title=\centering\textbf{Soft Prompt Design in LGSPF}, 
        arc=1.5mm, 
        boxrule=0.5pt, 
        left=2.5mm, 
        right=2.5mm, 
        top=1.5mm, 
        bottom=1.5mm
    ]
    \small
    \linespread{1.2}\selectfont 
    
    $\mathbf{T}_{\text{inst}}$: You are a fraud detection expert analyzing user behavior on a multi-relational graph dataset. We inject relation-specific structural representations into the hybrid template as follows:
    
    \vspace{1.5mm} 
    \textbf{\textcolor{blue!60!gray}{\texttt{<|$s_1$|>}}}: $\mathbf{T}_{\text{desc}_1}$ (Introduction to the Relation 1).  \\
    \textbf{\textcolor{blue!60!gray}{\texttt{<|$s_2$|>}}}: $\mathbf{T}_{\text{desc}_2}$ (Introduction to the Relation 2). \\
    \textbf{\textcolor{blue!60!gray}{\texttt{<|$s_3$|>}}}: $\mathbf{T}_{\text{desc}_3}$ (Introduction to the Relation 3).
    
    \vspace{1.5mm} 
    $\mathbf{T}_{\text{quest}}$: 
    \texttt{Q}: Is this node fraudulent or normal? You must analyze the multi-relational soft tokens and respond with one of these two words: 'fraud' or 'normal'. \texttt{A}:

    \vspace{1.5mm} 
    \tcbox[on line, size=fbox, boxrule=0.5pt, arc=1mm, colback=teal!5!white, colframe=teal!50!gray, left=1.5mm, right=1.5mm, top=1mm, bottom=1mm]{%
        \textbf{\textcolor{teal!70!black}{Target Generation:}} \texttt{fraud} $\vert$ \texttt{normal}%
    }
    \end{tcolorbox}
    \caption{The formal blueprint of the hybrid natural language template designed in LGSPF. The explicitly highlighted target space indicates the expected completion during the instruction tuning phase.}
    \label{box:prompt_design}
\end{table}

\subsection{Soft Prompt and Semantic Mapping}

In natural language processing, soft prompt \citep{lester2021power} is considered as a set of learnable continuous vectors, which does not match exact tokens in the vocabulary. Inspired by this, we treat the topological representations extracted by GNNs as conditional soft prompt. This design enables an alignment of the graph's structural characteristics with the latent space of the LLMs, preventing the loss of high-dimensional continuous information.

As explicitly exemplified in Table~\ref{box:prompt_design}, we denote the vocabulary of the LLM as $\mathcal{T} = \{t_1, \dots, t_{|\mathcal{T}|}\}$. The token embedding matrix is defined as $\mathbf{W}_{e} \in \mathbb{R}^{{|\mathcal{T}|} \times d_{\text{emb}}}$, where $d_{\text{emb}}$ is the embedding dimension. Specifically, given a discrete input token
$t_i \in \mathcal{T}$ (where $i \in \{1, \dots, |\mathcal{T}|\}$), 
the embedding layer applies a mapping function $f_{\text{emb}}(\cdot)$ to extract its corresponding continuous representation by retrieving the $i$-th row of the embedding matrix, denoted as
\begin{equation}
    \label{eq:3}
    \mathbf{e}_{t_i} = f_{\text{emb}}(t_i) = 
    \mathbf{W}_{e}[i, :] \in 
    \mathbb{R}^{d_{\text{emb}}}.
\end{equation}

In order to help LLMs distinguish the semantic difference between relations, we construct a hybrid prompt template $\mathcal{P}_{\text{temp}}$, which consists of natural language instructions and special tokens in an alternating manner :
\begin{equation}
\label{eq:8}
\begin{split}
\mathcal{P}_{\text{temp}} = [ \mathbf{T}_{\text{inst}}, s_1, \mathbf{T}_{\text{desc}_1}, \dots, \\ s_m, \mathbf{T}_{\text{desc}_m}, \mathbf{T}_{\text{quest}} ]
\end{split}
\end{equation}
where the $\mathbf{T}_{\text{inst}}$ is defined as the token sequence of the global task instructions while the $\mathbf{T}_{\text{quest}}$ denotes the instructions of classification tasks. We realize the semantic mapping through the pairing of special token $s_j \in \mathcal{S}$ which represents the topological features under subgraphs $\mathcal{G}_{r_j}$ and the descriptions  $\mathbf{T}_{\text{desc}_j}$ of relations ${r_j}$ in natural language.

To accommodate the special tokens from $s_1$ to $s_m$, we enlarge the original vocabulary $\mathcal{T}$ to an augmented token space $\mathcal{T}' = \mathcal{T} \cup \mathcal{S}$ where $\mathcal{S} = \{ s_1 ,\dots, s_m \}$. Consequently, the entire hybrid sequence $\mathcal{P}_{\text{temp}}$ is first mapped into an initial embedding matrix $\mathbf{E}_{\text{temp}}$ via the pre-trained embedding layer $f_{\text{emb}}(\cdot)$:
\begin{equation}
   \label{eq:9}
   \mathbf{E}_{\text{temp}} = f_{\text{emb}}(\mathcal{P}_{\text{temp}}) \in \mathbb{R}^{L \times d_{\text{emb}}}
\end{equation}
where $L$ denotes the total counts of token nums in the conditional soft prompt sequence. 

Finally, to dynamically incorporate structural features of the target node under multiple relations, we perform a structure-aware injection, replacing the initial embeddings of the special tokens in $\mathbf{E}_{\text{temp}}$ with the relation-specific structural embeddings $\mathbf{h}_{v_i, r_j}$ extracted from the parallel GNN encoders. Let $\text{idx}(s_j)$ be the absolute position index of the special tokens $s_j$ within the sequence $\mathcal{P}_{\text{temp}}$. The final input embedding matrix $\mathbf{E}_{\text{input}}$ is constructed as follows:
\begin{equation}
   \label{eq:10}
   \mathbf{E}_{\text{input}}[k, :] = 
   \begin{cases} 
   \mathbf{h}_{v_i, r_j}, & \text{if } k = \text{idx}(s_j), \\[2pt]
   & \forall j \in [1, m] \\[8pt]
   \mathbf{E}_{\text{temp}}[k, :], & \text{otherwise}
   \end{cases}
\end{equation}
Through this explicit cross-modal transition, the discrete special tokens are dynamically activated by continuous structural tensors, guiding the LLM to perform cognitive reasoning over complex multi-relational topologies.

\subsection{End-to-End Optimization}

We reframe the graph fraud detection task as a conditional text generation problem under a standard Causal Language Modeling paradigm. Specifically, we establish a deterministic mapping between the binary label space and the target text sequences: 
\begin{equation}
   \label{eq:11}
   \mathbf{Y}^{(y_i)} = 
   \begin{cases}
   \text{``fraud''}, & \text{if } y_i = 1 \\
   \text{``normal''}, & \text{if } y_i = 0
   \end{cases}
\end{equation}
To account for tokenizer variability across different LLM backbones, each target sequence is represented as a list of discrete tokens, i.e., $\mathbf{Y}^{(k)} = \{t^{(k)}_1, \dots, t^{(k)}_{N_k}\}$ where $k \in \{0, 1\}$. The conditional generation probability for each candidate sequence $\mathbf{Y}^{(k)}$ is computed via the chain rule:
\begin{equation}
   \label{eq:12}
   P(\mathbf{Y}^{(k)} \mid \mathbf{E}_{\text{input}}) = \prod_{j=1}^{N_k} P(t^{(k)}_j \mid \mathbf{E}_{\text{input}}, t^{(k)}_{<j})
\end{equation}
During the inference phase, the model computes the sequence-level log-likelihood for both candidate answers and generates the final predicted label:
\begin{equation}
   \label{eq:13}
   \hat{y}_i = \mathop{\arg\max}_{k \in \{0, 1\}} \log P(\mathbf{Y}^{(k)} \mid \mathbf{E}_{\text{input}})
\end{equation}

Our training goal is to minimize the discrepancy between the LLM's generated output and the ground-truth label text. We mask the labels for all prompt tokens, ensuring the loss is only computed on the answer tokens as follows:
\begin{equation}
  \label{eq:14}
  \mathcal{L} = - \sum_{j=1}^{N_{y_i}} \log P(t^{(y_i)}_j \mid \mathbf{E}_{\text{input}}, t^{(y_i)}_{<j})
\end{equation}
To reduce computational costs during training, we apply Low-Rank Adaptation \citep{hu2022lora} exclusively to the attention layers of the LLM. Formally, the complete set of trainable parameters $\Psi$ in LGSPF is formulated as:
\begin{equation}
  \label{eq:15}
  \Psi = \left\{ \bm{\Theta}_{r_j} \right\}_{j=1}^{m} \cup \left\{ \mathbf{A}_{l}, \mathbf{B}_{l} \right\}_{l \in \mathcal{L}_{\text{LoRA}}}
\end{equation}
where $\{\mathbf{A}_l, \mathbf{B}_l\}$ are the low-rank adapters in the $l$-th Transformer layer. Such an end-to-end optimization allows the LLM's semantic feedback to directly supervise the structural feature extraction. Consequently, the GNN encoders can adaptively calibrate their topological representations to better align with the LLM's cognitive space, leading to a deeper structural-semantic convergence.

\section{Experiments}

\subsection{Experimental Setup}

\paragraph{Datasets} To evaluate the effectiveness of our proposed LGSPF, we conducted experiments on three real-world public fraud detection datasets. The Amazon dataset \citep{mcauley2013amateurs} contains a selection of reviews on products under the Musical Instruments category. The YelpChi dataset \citep{rayana2015collective} includes hotel and restaurant reviews on the Yelp platform. The S-FFSD is a simulated graph dataset from a financial fraud semi-supervised situation \citep{xiang2023semi}. For the Amazon and YelpChi dataset, we process them into multi-relational graph following the same steps in \citet{dou2020enhancing}. The key properties of the datasets are presented in Table~\ref{tab:dataset_properties}.
Detailed descriptions to the experimental datasets are provided in Appendix~\ref{sec:appendix_datasets}.

\begin{table}
  \centering
  \resizebox{\columnwidth}{!}{%
  \begin{tabular}{c c c c c c}
    \hline
    \textbf{Dataset} & \textbf{Nodes} & \textbf{Fraud} & \textbf{Labeled} & \textbf{Relation} & \textbf{Edges} \\
    \hline
    Amazon & 11,944 & 6.87\% & 72.33\% & 
    \begin{tabular}{@{}c@{}} U-P-U \\ U-S-U \\ U-V-U \end{tabular} & 
    \begin{tabular}{@{}c@{}} 175,608 \\ 3,566,479 \\ 1,036,737 \end{tabular} \\
    \hline
    YelpChi & 45,954 & 14.53\% & 100\% & 
    \begin{tabular}{@{}c@{}} R-T-R \\ R-U-R \\ R-S-R \end{tabular} & 
    \begin{tabular}{@{}c@{}} 573,616 \\ 49,315 \\ 3,402,743 \end{tabular} \\
    \hline
    S-FFSD & 77,881 & 6.75\% & 38.06\% & 
    \begin{tabular}{@{}c@{}} Source \\ Target \\ Location \\ Type \end{tabular} & 
    \begin{tabular}{@{}c@{}} 120,224 \\ 229,170 \\ 232,275 \\ 232,721 \end{tabular} \\
    \hline
  \end{tabular}%
  }
  \caption{Key properties of the evaluated datasets. To emphasize the Weak-TAG scenario, the node features in all three datasets are numeric.}
  \label{tab:dataset_properties}
  \vspace{-5pt}
\end{table}

\paragraph{Baselines} We compared our LGSPF with a wide range of competitive models, which can be divided into two categories: (i) Classical GNNs, including GCN \citep{kipf2017semi}, GAT \citep{velivckovic2018graph}, GraphSAGE \citep{hamilton2017inductive},CARE-GNN \citep{dou2020enhancing}, PC-GNN \citep{liu2021pick}, BWGNN \citep{tang2022rethinking}, GTAN \citep{xiang2023semi} and RGTAN \citep{xiang2025enhancing}. We implemented these models using official code; (ii) Text-flattened LLMs, which directly serialize numerical features of the target node and its neighbor into discrete textual sequences, including Zero-shot LLM relying purely on pre-trained models and Instruct LLM fine-tuned via LoRA. Detailed prompt designs for them could be found in Appendix~\ref{sec:appendix_prompts}.

\paragraph{Implementation Settings}  For all LLM-related models in our experiments, we employ the Qwen2.5-7B-Instruct \citep{qwen2.5} as the LLM backbone and apply LoRA \citep{hu2022lora} to the attention layers of parameter-efficient fine-tuning. For the GNN component, we set the neighborhood aggregation depth to $K = 2$. We randomly split all three datasets into training, validation, and testing sets with a ratio of 40\%, 20\%, and 40\%. The end-to-end optimization is realized using the Adam optimizer \citep{kingma2014adam} with a learning rate of $3 \times 10^{-4}$ across all three datasets. We conduct our experiments on a Linux system with 4 NVIDIA A800 GPUs (80GB memory each).

\paragraph{Evaluation Metrics} We evaluate the experimental results over the three datasets under the ROC curve (AUC), Recall and Geometric Mean (G-Mean). LLMs generate outputs autoregressively based on token logits and sampling strategies. To compute the AUC, we define an anomaly score $\mathcal{S}_i$ for each node $v_i$. This score is derived from the log-likelihood margin between the fraudulent and normal target sequences:
\begin{equation}
   \label{eq:auc_score}
   \mathcal{S}_i = \log P(\mathbf{Y}^{(1)} \mid \mathbf{E}_{\text{input}}) - \log P(\mathbf{Y}^{(0)} \mid \mathbf{E}_{\text{input}})
\end{equation}
This formulation naturally accommodates multi-token sequences caused by diverse tokenization patterns and prompt designs. We evaluate Recall and G-Mean based on the final generated label $\hat{y}_i$.

\begin{table*}[t]
    \centering
    \resizebox{\textwidth}{!}{%
    \begin{tabular}{lccccccccc}
        \toprule
        \textbf{Method} & \multicolumn{3}{c}{\textbf{Amazon}} & \multicolumn{3}{c}{\textbf{YelpChi}} & \multicolumn{3}{c}{\textbf{S-FFSD}} \\
        \cmidrule(lr){2-4} \cmidrule(lr){5-7} \cmidrule(lr){8-10}
        & \textbf{AUC} & \textbf{Recall} & \textbf{G-Mean} & \textbf{AUC} & \textbf{Recall} & \textbf{G-Mean} & \textbf{AUC} & \textbf{Recall} & \textbf{G-Mean} \\
        \midrule
        GCN        & 0.8362 & 0.7751 & 0.7723 & 0.6338 & 0.5906 & 0.5931 & 0.7404 & 0.5777 & 0.6824 \\
        GAT & 0.7954 & 0.7234 & 0.7296 & 0.5704 & 0.5363 & 0.5429 & 0.7051 & 0.5611 & 0.6583 \\
        GraphSAGE  & 0.8735 & 0.7842 & 0.8211 & 0.7374 & 0.6325 & 0.6146 & 0.6713 & 0.3814 & 0.6055 \\
        CARE-GNN & 0.9200 & 0.8635 & 0.8824 & 0.7727 & 0.7046 & 0.7044 & 0.7112 & 0.4870 & 0.6230 \\
        PC-GNN       & 0.9665 & 0.9018 & 0.9062 & 0.8184 & 0.7348 & 0.7290 & 0.7033 & 0.5705 & 0.6213 \\
        BWGNN & \underline{0.9717} & 0.7872 & 0.8840 & 0.8876 & 0.6579 & 0.7695 & 0.7029 & 0.6201 & 0.6199 \\
        GTAN      & 0.9585 & 0.7561 & 0.8680 & 0.9172 & 0.5600 & 0.7363 & 0.8299 & 0.6921 & 0.6196 \\
        RGTAN     & 0.9638 & 0.8898 & 0.8836 & \underline{0.9194} & 0.7743 & 0.7506 & 0.8246 & 0.6940 & 0.6279 \\
        Zero-shot LLM                   & 0.5062 & 0.0321 & 0.1453 & 0.4145 & 0.0172 & 0.0284 & 0.4633 & 0.0304 & 0.1729 \\
        Instruct-LLM                    & 0.8390 & 0.9077 & 0.7665 & 0.7124 & 0.6710 & 0.7444 & 0.7188 & 0.4408 & 0.6412 \\
        \midrule
        w/o LLM                         & 0.9087 & 0.7812 & 0.8812 & 0.7726 & 0.5947 & 0.7395 & 0.7638 & 0.4912 & 0.6877 \\
        w/o Semantics                   & 0.9640 & 0.8863 & 0.9197 & 0.9095 & 0.7775 & 0.6418 & 0.8364 & 0.7432 & \underline{0.7801} \\
        w/o Joint-Opt                   & 0.9703 & \textbf{0.9271} & \underline{0.9233} & 0.9076 & \underline{0.7915} & \underline{0.8296} & \underline{0.8410} & \underline{0.7476} & 0.7588 \\
        \midrule
        \textbf{LGSPF} & \textbf{0.9805} & \underline{0.9179} & \textbf{0.9310} & \textbf{0.9208} & \textbf{0.8346} & \textbf{0.8325} & \textbf{0.8719} & \textbf{0.7561} & \textbf{0.7837} \\
        \bottomrule
    \end{tabular}%
    }
    \caption{Overall performance comparison and ablation studies across three benchmark datasets. All reported results are the averages of $5$ independent runs under the same random seeds.}
    \label{tab:main_results}
\end{table*}

\subsection{Overall Evaluation}

Table~\ref{tab:main_results} summarizes the overall performance of all compared methods. Compared to classical general-purpose graph models including GCN, GAT, and GraphSAGE, the models specifically designed for graph fraud detection such as CARE-GNN, PC-GNN, BWGNN, and GTAN exhibit consistently superior performance across all three datasets. Since fraudulent targets often show structural camouflage and heterogeneous relational patterns, fraud-specific methods overcome these challenges from perspectives such as noise filtering and class distribution balancing. Our experimental results confirm their effectiveness and improvements.
The Text-flattened LLMs exhibit severely degraded performance, unexpectedly lagging behind even the classical GNN baselines. This failure explicitly exposes the inherent semantic dilemma LLMs face in Weak-TAG graphs: directly serializing high-dimensional floating-point arrays into discrete text prompts forces standard tokenizers to fragment continuous structural features.

Our proposed LGSPF achieves superior performance, consistently outperforming all baselines by a significant margin across the three datasets. LGSPF elegantly resolves the aforementioned semantic dilemma. By mapping GNN-extracted topologies into continuous soft prompts and bridging them with hybrid natural language templates, our framework successfully aligns structural features with the LLM's cognitive space without information degradation. While traditional task-specific GNNs struggle to explicitly interpret the high-order semantic differences among complex multi-relational topologies, LGSPF overcomes this bottleneck by leveraging the LLM's deep semantic comprehension, ultimately leading to a comprehensive and robust enhancement in graph fraud detection tasks.

\subsection{Ablation Studies}

In Table~\ref{tab:main_results}, we also conduct three ablation experiments to evaluate the effectiveness of the key components and designs in LGSPF. To validate the impact of the semantic reasoning and soft prompt, we introduce two ablation variants. In w/o LLM, we substitute the LLM backbone with a simple MLP classifier, where the outputs from the parallel GNN encoders are directly concatenated. In w/o Semantics, we remove all natural language instructions and relational descriptions included in Eq.~\ref{eq:8}, leaving only the isolated continuous graph tokens $s_1, \dots, s_m$. Both variants result in a clear performance drop, which proves that the LLM possesses superior discriminative capacity for multi-relational representations. Furthermore, the textual descriptions serve as vital semantic anchors, effectively unlocking the LLM's potential to utilize its pre-trained real-world knowledge for enhanced fraud identification. To validate the necessity of end-to-end optimization, we design w/o Joint-Opt by cutting off the gradient backpropagation chain at the LLM-GNN interface. Specifically, we solely fine-tune the LLM while keeping the parallel GNNs frozen with random initialization. Although the performance slightly declines, it achieves several sub-optimal results across all three datasets. This phenomenon highlights the strong fine-grained semantic comprehension of LLMs, which can detect fraud patterns even with non-calibrated structural embeddings. Nevertheless, the end-to-end optimization promotes a deeper cross-modal alignment, elevating the overall performance.

\begin{figure}
  \includegraphics[width=\columnwidth]{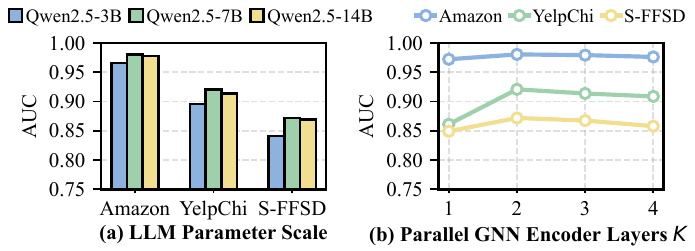}
  \caption{Sensitivity analysis of the LLM parameter scale and the aggregation depth $K$ across three benchmark datasets evaluated by AUC.}
  \label{fig:sensitivity}
\end{figure}

\subsection {Sensitivity Studies}

In this section, we study the parameter sensitivity of our proposed LGSPF by varying the scale of the LLM backbone and the neighborhood aggregation depth $K$ for the parallel GNN encoders. The experimental results across the three detection datasets are reported in Figure~\ref{fig:sensitivity}. Specifically, we vary the parameter scale of the Qwen2.5 series backbone among \{3B, 7B, 14B\} and adjust the depth $K$ from 1 to 4.
Regarding the capacity of the LLM backbone, we observe a significant performance improvement when upgrading the scale from 3B to 7B, while further scaling the LLM to 14B yields saturation or slight degradation. This indicates that a threshold of parameter scale is essential to comprehend complex structural semantics; however, it also implies that massive models might suffer from slight overfitting during parameter-efficient fine-tuning. Ultimately, the 7B model strikes the optimal balance between reasoning capability and adaptation efficiency, making it the most cost-effective choice for our framework. Regarding the structural receptive field, performance peaks at $K=2$ and declines for $K \ge 3$. While $K=1$ restricts the receptive field and misses higher-order topological fraud patterns, excessive depths ($K \ge 3$) invariably encounter the over-smoothing issue. Therefore, a moderate depth ($K=2$) is essential to balance sufficient topological aggregation with the sharp discriminability of fraudulent nodes.

\begin{table}
    \centering
    \resizebox{\columnwidth}{!}{
    \begin{tabular}{l ccc}
        \toprule
        \textbf{Embedded View} & \textbf{AUC} & \textbf{Recall} & \textbf{G-Mean} \\
        \midrule
        Single-view R-S-R & 0.8680 & 0.7043 & 0.7620 \\
        Single-view R-T-R & 0.8531 & 0.8125 & 0.7187 \\
        Single-view R-U-R & 0.8927 & 0.7504 & 0.8049 \\
        \midrule
        Single-view Average* & 0.8713 & 0.7557 & 0.7619 \\
        \textbf{Full view (LGSPF)} & \textbf{0.9208} & \textbf{0.8346} & \textbf{0.8325} \\
        \bottomrule
    \end{tabular}
    }
    \caption{Interpretability study on the YelpChi dataset. The Single-view Average* represents the arithmetic mean of the three individual relation metrics.}
    \label{tab:interpretability}
    \vspace{-5pt}
\end{table}

\subsection{Interpretability Studies}

To explore the potential of LGSPF in advancing the interpretability of Graph Fraud Detection, we conduct a single-relation embedding experiment on the YelpChi dataset. Specifically, we degenerate the comprehensive multi-relational prompt sequence by replacing the entire relation block $\{s_k, \mathbf{T}_{\text{desc}_k}\}_{k=1}^m$ with a single embedding-description pair $\{s_j, \mathbf{T}_{\text{desc}_j}\}$. As reported in Table~\ref{tab:interpretability}, the detection performance fluctuates across different isolated relations. For instance, the R-U-R view achieves the highest AUC among three single views and the R-T-R view exhibits an advantage in Recall. This result demonstrates that distinct relational connections imply differentiated structural semantics. However, the integrated Full view decisively outperforms the Single-view Average* by 5.4\%, 10.4\%, and 8.5\% in terms of AUC, Recall, and G-Mean, respectively. By prompting the LLM to process diverse relations simultaneously, LGSPF effectively mines the intrinsic correlations among multiple relations, achieving a deeper cross-semantic fusion that boosts detection performance. Furthermore, the high modularity of LGSPF allows for dynamic adjustments regarding the type and number of injected views based on domain-specific knowledge. This flexibility provides a powerful tool for diagnosing fraudulent mechanisms in different application contexts, shedding new light on advancing graph fraud detection from black-box predictions to behavior recognition.

\section{Conclusion}

In this paper, we propose LGSPF, a novel LLM-GNN Soft Prompt Framework for fraud detection. To overcome the challenge of deploying LLMs in Weak Text-Attributed Graph (Weak-TAG) scenarios where rich textual data is unavailable, we implement a continuous soft prompting mechanism that bridges the graph topology with the LLM's semantic space, successfully bypassing the feature distortion caused by discrete textualization. Furthermore, to address the multi-relational complexity inherent in fraudulent activities and the structural-semantic misalignment, we design a parallel GNN encoder alongside an end-to-end optimization paradigm. This architecture empowers the LLM to dynamically reason over distinct relational topologies. Extensive experiments across multiple benchmarks demonstrate that LGSPF achieves state-of-the-art performance. Beyond empirical superiority, our framework significantly enhances the semantic interpretability of fraud behaviors, providing a highly promising paradigm for advancing fraud detection from black-box predictions to transparent behavior recognition in privacy-sensitive domains.

\section*{Limitations}

Although our proposed LGSPF framework demonstrates remarkable performance in multi-relational Graph Fraud Detection under the Weak-TAG setting, several limitations remain to be addressed in future research. First, regarding scalability, while we employ Parameter-Efficient Fine-Tuning and lightweight parallel GNN encoders to minimize resource consumption, the integration of Large Language Models inevitably introduces substantial computational overhead. Consequently, directly deploying LGSPF on ultra-large-scale industrial graph datasets with billions of nodes and edges remains computationally demanding, particularly in resource-constrained environments. Secondly, the framework's performance intrinsically depends on the cognitive reasoning capabilities of the pre-trained LLM backbone. Due to hardware limitations, our main experiments are conducted using Qwen2.5-7B model. This constraint may prevent the framework from fully unlocking and deeply mining more complex structural semantics. In the future, we aim to validate our framework on larger-scale foundational models to explore performance across more diverse fraud detection scenarios. Finally, the current implementation of LGSPF is primarily tailored for static multi-relational graphs. In real-world applications, fraudulent nodes and interactive behaviors frequently evolve over time. The absence of temporal dynamics restricts the model from capturing the chronological progression of camouflage strategies. In future work, we plan to extend our framework to dynamic graphs by incorporating temporal embeddings, thereby further improving its effectiveness and practical applicability.

\bibliography{custom}

\appendix
\section{Datasets and Multi-Relational Construction}
\label{sec:appendix_datasets}

To guarantee the reproducibility of our experimental outcomes, we provide a thorough description regarding the background, raw node feature configuration, and multi-relational graph instantiation for the three benchmark datasets used in this study. Following the preprocessing paradigms outlined by \citet{dou2020enhancing} and \citet{xiang2023semi}, all unstructured attributes are discarded to enforce the Weak-TAG setting. The hand-crafted input feature dimensions for the Amazon, YelpChi, and S-FFSD datasets are rigorously specified as 25, 32, and 126, respectively.

\subsection{Amazon Dataset}
The Amazon dataset \citep{mcauley2013amateurs} operates within the e-commerce domain, capturing user review dynamics centered around product entities under the \textit{Musical Instruments} category. Following established conventions, users maintaining greater than 80\% helpful upvotes are designated as benign entities ($y=0$), while individuals with fewer than 20\% helpful votes are categorized as fraudulent accounts ($y=1$).  We construct a multi-relational graph structure $\mathcal{G} = (\mathcal{V}, \mathcal{E}, \mathcal{R}_A)$ governed by three specific relationship types: $\mathcal{R}_A = \{r_{A1}, r_{A2}, r_{A3}\}$.
\begin{itemize}
    \item \textbf{U-P-U Relation ($r_{A1}$):} Connects two reviewer nodes if they have authored a review for the identical product entity. This relation serves to isolate co-review target syndication.
    \item \textbf{U-S-U Relation ($r_{A2}$):} Connects two reviewer nodes if they assigned an identical star rating to any product within a tightly bounded temporal window of one week. It captures synchronized rating distortion.
    \item \textbf{U-V-U Relation ($r_{A3}$):} Connects two reviewer nodes if their textual review similarities reside within the top 5\% of the global distribution, where similarity is computed via cosine distance over hand-crafted TF-IDF vectors.
\end{itemize}

\subsection{YelpChi Dataset}
The YelpChi dataset \citep{rayana2015collective} focuses on crowdsourced business intelligence, encompassing hotel and restaurant reviews processed by Yelp's proprietary anti-fraud filter. Review entities recommended as legitimate are denoted as normal ($y=0$), whereas filtered reviews are tagged as spam/fraudulent ($y=1$). Nodes are represented by 32-dimensional hand-crafted statistical features. We construct a multi-relational graph structure $\mathcal{G} = (\mathcal{V}, \mathcal{E}, \mathcal{R}_Y)$ governed by three specific relationship types: $\mathcal{R}_Y = \{r_{Y1}, r_{Y2}, r_{Y3}\}$.
\begin{itemize}
    \item \textbf{R-T-R Relation ($r_{Y1}$):} Connects two review nodes if they correspond to the exact same business establishment and were posted within the identical calendar month, uncovering burst-mode review injection.
    \item \textbf{R-U-R Relation ($r_{Y2}$):} Connects two review nodes if they were generated by the same unique user account, tracking persistent fraudulent identity footprints.
    \item \textbf{R-S-R Relation ($r_{Y3}$):} Connects two review nodes if they share the same target business and hold identical star ratings, effectively capturing coordinated sentiment manipulation clusters.
\end{itemize}

\subsection{S-FFSD Dataset}
The Simulated Financial Fraud Semi-Supervised Dataset (S-FFSD) is an open-source, simulated, and scaled-down variant of the original proprietary FFSD dataset introduced by \citet{xiang2023semi}. It simulates an anonymous credit card transactional ecosystem where nodes denote individual financial transactions, and the numerical features are structured into a 126-dimensional dense vector space. Ground-truth labels are authenticated via real-world consumer reporting: transactions verified as fraudulent are marked as positive ($y=1$), normal transactions as negative ($y=0$), and unverified records are masked as unlabelled ($y=2$). 

Distinct from standard undirected graphs, the topological connections in S-FFSD are strictly governed by temporal dynamics. We decode the raw attributes \textit{Source}, \textit{Target}, \textit{Location}, and \textit{Type} to partition transactions into discrete groups. Within each specific group, nodes are sorted chronologically by their execution timestamps. Directed edges are subsequently constructed pointing from earlier transactions to later ones. To capture localized temporal dependencies while avoiding the computational bottleneck caused by dense over-connections, each transaction is explicitly connected only to the subsequent $k=3$ transactions within the same group. Finally, self-loops are universally injected into the adjacency matrices to retain original node features during the GNN message-passing phase. 

Following this temporal-directed paradigm, we construct a multi-relational graph structure $\mathcal{G} = (\mathcal{V}, \mathcal{E}, \mathcal{R}_F)$ governed by four specific relationship types: $\mathcal{R}_F = \{r_{F1}, r_{F2}, r_{F3}, r_{F4}\}$.
\begin{itemize}
    \item \textbf{Source Relation ($r_{F1}$):} Connects two transaction nodes, directed from the earlier transaction to the later one, if they originate from the exact same sender entity, tracking high-frequency velocity attacks from single compromised accounts.
    \item \textbf{Target Relation ($r_{F2}$):} Connects two transaction nodes, directed from the earlier transaction to the later one, if they share the identical receiver entity, facilitating the detection of merchant-side fraud collection points.
    \item \textbf{Location Relation ($r_{F3}$):} Connects two transaction nodes, directed from the earlier transaction to the later one, if they are executed within the identical geographical or electronic location identifier, isolating location-based card-cloning anomalies.
    \item \textbf{Type Relation ($r_{F4}$):} Connects two transaction nodes, directed from the earlier transaction to the later one, if they possess matching commercial transaction type codes, contextualizing the behavioral domain of fraudulent capital outflows.
\end{itemize}

\section{Prompt Designs}
\label{sec:appendix_prompts}

In this section, we present the exact prompt templates utilized in our experiments. Figure~\ref{fig:LGSPF_prompts} details the multi-relational hybrid prompt structures employed by our proposed LGSPF framework across the three benchmark datasets. Figure~\ref{fig:baseline_prompts} shows the serialization templates used by the Text-flattened LLM baselines.

\begin{figure*}[htbp]
    \centering
    \begin{tcolorbox}[colback=blue!4!white, colframe=blue!50!gray, title=\textbf{LGSPF Prompt Template for Amazon Dataset}, arc=2mm, boxrule=0.5pt, left=2mm, right=2mm, top=1mm, bottom=1mm]
    \small
    You are a fraud detection expert analyzing user review behavior on a graph dataset. \\
    This dataset is a multi-relational graph dataset constructed from user reviews of Amazon website products. \\
    We apply the GraphSAGE algorithm to obtain node embeddings under different relational views as follows: \\
    \textbf{\textcolor{blue!80!black}{\texttt{<|graph\_pad\_relation1|>}}}: U-P-U relation embedding --- connects users who reviewed the same product. \\
    \textbf{\textcolor{blue!80!black}{\texttt{<|graph\_pad\_relation2|>}}}: U-S-U relation embedding --- connects users who gave the same star rating within a one-week window. \\
    \textbf{\textcolor{blue!80!black}{\texttt{<|graph\_pad\_relation3|>}}}: U-V-U relation embedding --- connects users with top 5\% mutual review text similarity based on TF-IDF. \\
    Q: Is this review node fraudulent or normal? \\
    You must analyze the embedding patterns and respond with exactly one of these two words: 'fraud' or 'normal' \\
    A: \texttt{[fraud / normal]}
    \end{tcolorbox}
    
    \vspace{2mm}
    
    \begin{tcolorbox}[colback=blue!4!white, colframe=blue!50!gray, title=\textbf{LGSPF Prompt Template for YelpChi Dataset}, arc=2mm, boxrule=0.5pt, left=2mm, right=2mm, top=1mm, bottom=1mm]
    \small
    You are a fraud detection expert analyzing user review behavior on a graph dataset. \\
    This dataset is a multi-relational graph dataset constructed from user reviews of hotels and restaurants on the Yelp platform. \\
    We apply the GraphSAGE algorithm to obtain node embeddings under different relational views as follows: \\
    \textbf{\textcolor{blue!80!black}{\texttt{<|graph\_pad\_relation1|>}}}: R-T-R relation embedding --- connects reviews posted on the same business within the same month, revealing temporal co-review behaviors. \\
    \textbf{\textcolor{blue!80!black}{\texttt{<|graph\_pad\_relation2|>}}}: R-U-R relation embedding --- connects reviews written by the same user, capturing user-level review consistency and behavioral tendencies. \\
    \textbf{\textcolor{blue!80!black}{\texttt{<|graph\_pad\_relation3|>}}}: R-S-R relation embedding --- connects reviews with the same star rating for the same business, reflecting rating alignment or potential opinion manipulation. \\
    Q: Is this review node fraudulent or normal? \\
    You must analyze the embedding patterns and respond with exactly one of these two words: 'fraud' or 'normal' \\
    A: \texttt{[fraud / normal]}
    \end{tcolorbox}
    
    \vspace{2mm}

    \begin{tcolorbox}[colback=blue!4!white, colframe=blue!50!gray, title=\textbf{LGSPF Prompt Template for S-FFSD Dataset}, arc=2mm, boxrule=0.5pt, left=2mm, right=2mm, top=1mm, bottom=1mm]
    \small
    You are a fraud detection expert analyzing transaction behavior on a graph dataset. \\
    This dataset is a multi-relational graph dataset constructed from financial transaction records in the FFSD (Financial Fraud Semi-supervised Dataset). \\
    We apply the GraphSAGE algorithm to obtain node embeddings under different relational views as follows: \\
    \textbf{\textcolor{blue!80!black}{\texttt{<|graph\_pad\_relation1|>}}}: Source relation embedding --- connects each transaction to the next 3 transactions with the same Source, ordered by time. \\
    \textbf{\textcolor{blue!80!black}{\texttt{<|graph\_pad\_relation2|>}}}: Target relation embedding --- connects each transaction to the next 3 transactions with the same Target, ordered by time. \\
    \textbf{\textcolor{blue!80!black}{\texttt{<|graph\_pad\_relation3|>}}}: Location relation embedding --- connects each transaction to the next 3 transactions with the same Location, ordered by time. \\
    \textbf{\textcolor{blue!80!black}{\texttt{<|graph\_pad\_relation4|>}}}: Type relation embedding --- connects each transaction to the next 3 transactions with the same Type, ordered by time. \\
    Q: Is this transaction node fraudulent or normal? \\
    You must analyze the embedding patterns and respond with exactly one of these two words: 'fraud' or 'normal' \\
    A: \texttt{[fraud / normal]}
    \end{tcolorbox}
    \caption{The hybrid natural language prompt templates implemented in LGSPF. The explicitly highlighted special tokens (e.g., \textbf{\textcolor{blue!80!black}{\texttt{<|graph\_pad\_relationX|>}}}) serve as the cross-modal interfaces, which are dynamically replaced by continuous relation-specific structural embeddings output by the parallel GNN encoders during the forward pass.}
    \label{fig:LGSPF_prompts}
\end{figure*}

\begin{figure*}[htbp]
    \centering
    \begin{tcolorbox}[colback=red!4!white, colframe=red!50!gray, title=\textbf{Text-flattened LLM Baseline Template for Amazon Dataset}, arc=2mm, boxrule=0.5pt, left=2mm, right=2mm, top=1mm, bottom=1mm]
    \small
    You are a fraud detection expert analyzing user review behavior on a graph dataset. \\
    This dataset is a multi-relational graph dataset constructed from user reviews of Amazon website products. \\
    We summarize the objective node features and its 1-hop neighbor features under different relational views as follows: \\
    \textbf{\textcolor{red!70!black}{\texttt{[[Flattened Target Node Features]]}}}: object node features. \\
    \textbf{\textcolor{red!70!black}{\texttt{[[Flattened Neighbor Features]]}}}: U-P-U relation features --- connects users who reviewed the same product. \\
    \textbf{\textcolor{red!70!black}{\texttt{[[Flattened Neighbor Features]]}}}: U-S-U relation features --- connects users who gave the same star rating within a one-week window. \\
    \textbf{\textcolor{red!70!black}{\texttt{[[Flattened Neighbor Features]]}}}: U-V-U relation features --- connects users with top 5\% mutual review text similarity based on TF-IDF. \\
    Q: Is this review node fraudulent or normal? \\
    You must analyze the embedding patterns and respond with exactly one of these two words: 'fraud' or 'normal' \\
    A: \texttt{[fraud / normal]}
    \end{tcolorbox}
    
    \vspace{2mm}
    
    \begin{tcolorbox}[colback=red!4!white, colframe=red!50!gray, title=\textbf{Text-flattened LLM Baseline Template for YelpChi Dataset}, arc=2mm, boxrule=0.5pt, left=2mm, right=2mm, top=1mm, bottom=1mm]
    \small
    You are a fraud detection expert analyzing user review behavior on a graph dataset. \\
    This dataset is a multi-relational graph dataset constructed from user reviews of hotels and restaurants on the Yelp platform. \\
    We summarize the objective node features and its 1-hop neighbor features under different relational views as follows: \\
    \textbf{\textcolor{red!70!black}{\texttt{[[Flattened Target Node Features]]}}}: object node features. \\
    \textbf{\textcolor{red!70!black}{\texttt{[[Flattened Neighbor Features]]}}}: R-T-R relation features --- connects reviews posted on the same business within the same month, revealing temporal co-review behaviors. \\
    \textbf{\textcolor{red!70!black}{\texttt{[[Flattened Neighbor Features]]}}}: R-U-R relation features --- connects reviews written by the same user, capturing user-level review consistency and behavioral tendencies. \\
    \textbf{\textcolor{red!70!black}{\texttt{[[Flattened Neighbor Features]]}}}: R-S-R relation features --- connects reviews with the same star rating for the same business, reflecting rating alignment or potential opinion manipulation. \\
    Q: Is this review node fraudulent or normal? \\
    You must analyze the embedding patterns and respond with exactly one of these two words: 'fraud' or 'normal' \\
    A: \texttt{[fraud / normal]}
    \end{tcolorbox}
    
    \vspace{2mm}

    \begin{tcolorbox}[colback=red!4!white, colframe=red!50!gray, title=\textbf{Text-flattened LLM Baseline Template for S-FFSD Dataset}, arc=2mm, boxrule=0.5pt, left=2mm, right=2mm, top=1mm, bottom=1mm]
    \small
    You are a fraud detection expert analyzing transaction behavior on a graph dataset. \\
    This dataset is a multi-relational graph dataset constructed from financial transaction records in the FFSD (Financial Fraud Semi-supervised Dataset). \\
    We summarize the objective node features and its 1-hop neighbor features under different relational views as follows: \\
    \textbf{\textcolor{red!70!black}{\texttt{[[Flattened Target Node Features]]}}}: object node features. \\
    \textbf{\textcolor{red!70!black}{\texttt{[[Flattened Neighbor Features]]}}}: Source relation features --- connects each transaction to the next 3 transactions with the same Source, ordered by time. \\
    \textbf{\textcolor{red!70!black}{\texttt{[[Flattened Neighbor Features]]}}}: Target relation features --- connects each transaction to the next 3 transactions with the same Target, ordered by time. \\
    \textbf{\textcolor{red!70!black}{\texttt{[[Flattened Neighbor Features]]}}}: Location relation features --- connects each transaction to the next 3 transactions with the same Location, ordered by time. \\
    \textbf{\textcolor{red!70!black}{\texttt{[[Flattened Neighbor Features]]}}}: Type relation features --- connects each transaction to the next 3 transactions with the same Type, ordered by time. \\
    Q: Is this transaction node fraudulent or normal? \\
    You must analyze the embedding patterns and respond with exactly one of these two words: 'fraud' or 'normal' \\
    A: \texttt{[fraud / normal]}
    \end{tcolorbox}
    \caption{The serialization prompt templates utilized by the Text-flattened LLM baselines. Contrast to LGSPF's continuous soft prompts, these baselines substitute the explicitly highlighted placeholders (\textbf{\textcolor{red!70!black}{\texttt{[[Flattened X Features]]}}}) directly with raw, discrete textual sequences composed of long arrays of numerical features, which induces semantic degradation and tokenization fragmentation during LLM reasoning.}
    \label{fig:baseline_prompts}
\end{figure*}

\end{document}